\documentclass[11pt]{article}

\usepackage[]{acl}
\usepackage{times}
\usepackage{latexsym}
\usepackage[T1]{fontenc}
\usepackage[utf8]{inputenc}
\usepackage{microtype}
\usepackage{inconsolata}
\usepackage{booktabs}
\usepackage{graphicx}
\usepackage{xurl}
\usepackage{adjustbox}
\usepackage{fancyvrb}

\title{When Better Turns Do Not Make Better Agents:\\
Diagnosing the Gap Between Next-Turn Metrics and Workflow Success\\
\texorpdfstring{%
  \raisebox{-0.2\height}{\includegraphics[height=7mm]{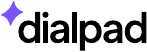}}%
}{}}

\author{Md Tahmid Rahman Laskar\thanks{Corresponding Author: tahmid.rahman@dialpad.com}, Xue-Yong Fu, \\ Gundeep Singh, Karol Chang, Kevin Sanders, Shi Zong, Tania Habib,\\ Julien Bouvier Tremblay, Shayna Gardiner, Harsh Saini, Matthias Lee, \\Elena Khasanova, Quinten McNamara, Shashi Bhushan TN \\ 
\textbf{(Bold in Author Names Denotes Equal Contribution)} \\ \textbf{Dialpad Inc.} }

\begin{document}
\maketitle

\begin{abstract}
Agent models are frequently evaluated one decision at a time, where the model predicts the next action based on the gold interaction history, which is scored against a reference. We investigate whether improvement under this protocol is predictive of improved autonomous workflow execution. We study pre-SFT and supervised fine-tuned (SFT) Qwen3 models at 4B and 14B parameters and Gemma 3 models at 4B and 12B parameters on multi-turn customer-support workflows. We find that SFT consistently improves text-turn success, and that overall next-turn success increases for every model under gold-history evaluation. However, these improvements do not transfer to autonomous workflow execution. Tool-specific gains also vary across metrics and models. None of the four SFT models succeeds under holistic workflow evaluation, with strict trajectory completion reaching at most 10.4\% workflow success. Our results show that next-turn evaluation is not a reliable proxy for workflow success, motivating separate reporting of text quality, local action correctness, tool execution, and end-to-end task completion.
\end{abstract}

\section{Introduction}

Language-model agents deployed in production settings must produce more than just plausible responses. For instance, a workflow agent needs to generate a factually correct answer, follow the domain policy, invoke tools with state-dependent arguments \citep{schick2023toolformer,qin2023toollm}, gather missing information, incorporate tool results, and recover from failed actions. An error early in an interaction changes the context for every later decision, so agent quality is fundamentally a property of a trajectory, not the average quality of its turns, a premise underlying recent interactive agent benchmarks \citep{yao2024taubench,barres2025tau2}. Nevertheless, next-turn evaluation, where the gold history precedes every decision and compares the model's prediction with a reference using a lexical or semantic similarity metric \citep{lin2004rouge,zhang2020bertscore,zha2023alignscore}, remains a dominant protocol in practice since it is cheap, reproducible, and verifiable. 

However, next-turn evaluation has two important limitations. First, because each prediction is based on the correct history, earlier mistakes cannot affect later decisions. It therefore measures how well a model responds from a correct state, rather than whether it can build and maintain that state during an interaction. Second, reference-based metrics may reward an incorrect response that resembles the reference while penalizing a valid response expressed differently. This concern has also been noted by \citet{alkhouli2025confetti} in their CONFETTI benchmark, in which they mention that the gold trajectory could ``artificially inflate'' later-turn performance through in-context learning. Nonetheless, they did not compare the turn-level evaluation in CONFETTI with end-to-end autonomous execution, where the model must continue from its own previous decisions. 


To this end, our study directly measures this gap using multi-turn customer-support workflows in the task-oriented dialogue evaluation setting \citep{qin2023end,budzianowski2018multiwoz}. Our research question is whether improvements in gold-history next-action scores predict improvements in end-to-end workflow success. We compare pre-SFT and supervised fine-tuned (SFT) \citep{wei2022finetuned,ouyang2022training} Qwen3 models \citep{yang2025qwen3} at 4B and 14B parameters and Gemma 3 models \citep{gemmateam2025gemma3} at 4B and 12B. Under gold-history evaluation, SFT improves text quality and overall turn-level performance across all model scales. However, when models must complete workflows using their own previous outputs, nearly all trajectories requiring tool use led to failure. Thus, the same SFT models that appear effective under turn-level evaluation are found to be ineffective in end-to-end workflow execution.

In this paper, we investigate three research questions : (i) whether turn-level improvements predict end-to-end workflow success, (ii) whether aggregate scores conceal differences between text generation and tool execution, and (iii) how early errors affect later decisions. Across multi-turn workflows, we find that SFT consistently improves gold-history turn-level performance but rarely enables successful execution of tool-requiring workflows. Separating response generation from tool use further shows that improvements in one capability can hide failures in the other, while trajectory analysis reveals error propagation that gold-history evaluation cannot capture. These findings motivate a minimal reporting framework that combines turn-level scores with end-to-end success and trajectory-level error analysis.
\section{Related Work}

\noindent \textbf{Text-generation and reference-based metrics.}
Reference-based metrics such as ROUGE measure lexical overlap \cite{lin2004rouge}, while BERTScore and AlignScore measure embedding similarity and source--candidate factual alignment, respectively \citep{zhang2020bertscore,zha2023alignscore}. These are well suited to isolated response turns but do not directly measure policy compliance or tool execution. LLM judges can recognize a valid response that differs from the reference, but their conclusions remain sensitive to the rubric and context supplied to the judge \citep{laskar2025judging,gu2026survey}.

\noindent \textbf{Tool-using language agents.}
A separate line of work studies models that select and invoke external tools, typically evaluating tool generalization across a broad catalog of functions \citep{schick2023toolformer,yao2023react,li2023apibank,qin2023toollm,laskar2026text}, with CONFETTI extending this to conversational, turn-level function calling \citep{alkhouli2025confetti}. Our setting instead fixes the tool catalog and governing policy per workflow in advance and asks whether an agent can execute the full multi-turn trajectory, including recovering from errors.

\noindent \textbf{Task-oriented dialogue and interactive agent benchmarks.}
Task-oriented dialogue has long studied goal completion through multi-turn interaction and API-grounded slot filling \citep{qin2023end,budzianowski2018multiwoz,rastogi2020schema}. Recent benchmarks extend evaluation to interactive environments \cite{liu2024agentbench}, including policy-constrained tool–agent–user interactions in $\tau$-bench and dual-control interaction in $\tau^2$-bench \cite{yao2024taubench,barres2025tau2}.

We complement these benchmarks by holding workflows and models fixed while varying only the evaluation protocol, isolating how gold-history, strict tool-matching, and closed-loop evaluation support different conclusions for identical outputs.

\noindent \textbf{Supervised fine-tuning for agent behavior.}
Supervised fine-tuning on instruction and dialogue data is a standard method for adapting models to follow instructions and use tools \citep{wei2022finetuned,ouyang2022training,wang2023selfinstruct,qu2025tool}, and compact open-weight models are an increasingly attractive target for this adaptation in cost-sensitive deployments \citep{fu2024tiny}. We treat SFT as a fixed, realistic intervention and ask whether its gains, measured under a gold-history protocol, transfer to closed-loop execution of the same workflows.
\begin{figure*}
    \centering
    \includegraphics[width=0.8\linewidth]{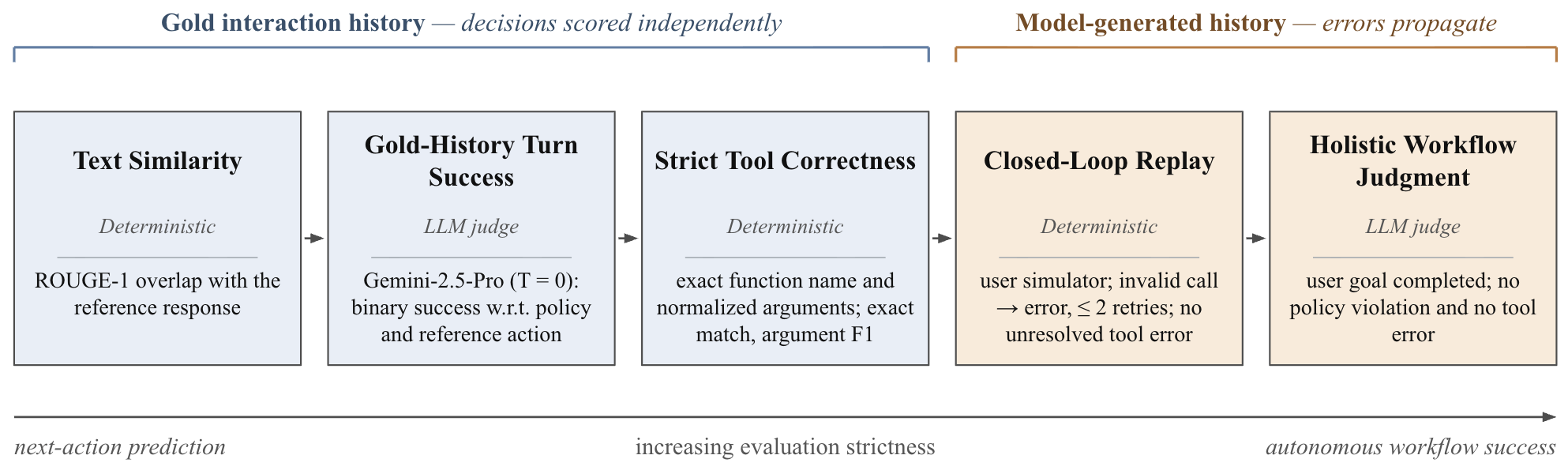}
    \caption{An Overview of our Evaluation Protocol}
    \label{fig:protocol}
\end{figure*}
\section{Experimental Setup}

\subsection{Data and Models}

\noindent \textbf{Training data.}
We collect a proprietary dataset from Dialpad\footnote{\url{https://www.dialpad.com/}} covering 130 customer-support workflows that are constructed from business conversations \cite{fu2022effective,laskar2023ai,khasanova2025dacip,laskar2025ai}. Each workflow is represented by the following: a domain policy, a set of available tool schemas, and one or more user goals. Following prior work on tool-use scenario generation and simulated users \citep{li2023apibank,qin2023toollm,yao2024taubench}, we construct scenarios that combine these elements and generate multi-turn conversations with two separate instances of GPT-5 \cite{singh2025openai}: one acts as the user and the other as the workflow agent. The resulting trajectories contain user messages, assistant responses, tool calls, and tool results.

At first, we generated 1800 conversations. Then, each generated conversation is independently checked by Claude-4.5-Opus\footnote{\url{https://www.anthropic.com/news/claude-opus-4-5}} and Gemini-2.5-Pro \cite{comanici2025gemini} for logical consistency, missing workflow steps, and policy compliance; a conversation is retained only when both judges approve it. This filtering removes 773 conversations and leaves 1,027 validated conversations. We then convert every assistant decision into a next-action example: the input contains the policy, tools, and preceding history, and the target is either a natural-language response or a structured tool call. The final training split contains 5,834 examples (4,093 text responses and 1,741 tool calls), with 664 additional validation examples (466 text and 198 tool). These generated conversations contain 13 turns on average. 

\noindent \textbf{Evaluation data.}
The held-out evaluation split contains 84 validated conversations, yielding 542 next-action examples: 376 natural-language responses and 166 tool calls. Conversations contain 2--18 assistant decisions (median 6). 
The split is held out at the conversation level, and no training conversation snippet is reused in evaluation. Gold-history metrics score the 542 examples independently; closed-loop evaluation instead assesses each model across the 84 complete conversations by leveraging their state-dependent tool results. 

\noindent \textbf{Models.}
We compare public Qwen3 models \citep{yang2025qwen3} at 4B and 14B parameters and Gemma 3 instruction-tuned models \citep{gemmateam2025gemma3} at 4B and 12B (\textsc{Pre-SFT}) with the corresponding models after full supervised fine-tuning (\textsc{SFT}). Every model uses the same training and validation examples and the same five-epoch recipe. 

\subsection{Evaluation Protocols}

We evaluate every model under five protocols, each targeting a different capability along the path from producing a plausible utterance to completing an autonomous workflow (see Figure \ref{fig:protocol}).

\noindent \textbf{(i) Text similarity.}
For every natural-language decision, the model receives the gold history preceding it, the policy, and the tool definitions, and we compare its response with the reference using the ROUGE-1 metric. 

\noindent \textbf{(ii) Gold-history turn success.}
Gemini-2.5-Pro at temperature zero judges all model-generated responses (542 turns, covering both text-based responses and tool calls) based on the given policy, tool definitions, gold history, and reference action. The judge model is required to return a binary success label by assessing action correctness while adhering to the domain policy. 

\noindent \textbf{(iii) Strict tool correctness.}
For the 166 reference tool-call decisions, we write a deterministic parsing script that requires an exact function-name and normalized-argument match between the prediction and reference. We report exact accuracy and argument F1, with no partial credit for a plausible but non-matching action.

\noindent \textbf{(iv) Closed-loop replay.}
We replay all 84 conversations using each model's own generated assistant history rather than gold history, with a deterministic user simulator supplying user turns. We use a deterministic parser to evaluate the tool calls. A correctly matched tool call receives the state-dependent result, while an invalid call receives an error and up to two retries. The workflow succeeds only if the model reaches the end of the conversation without any unresolved tool errors.  

\noindent \textbf{(v) Holistic workflow judgment.}
Gemini-2.5-Pro 
judges every closed-loop replay for full user-goal completion where successful workflows do not have any policy violation and no tool errors. 

\begin{table*}[t]
\centering
\small
\setlength{\tabcolsep}{3.5pt}
\begin{adjustbox}{max width=\textwidth}
\begin{tabular}{lllcccccccc}
\toprule
& & & \multicolumn{4}{c}{\textbf{Qwen3}} & \multicolumn{4}{c}{\textbf{Gemma 3}} \\
\cmidrule(lr){4-7}\cmidrule(lr){8-11}
& & & \multicolumn{2}{c}{\textbf{4B}} & \multicolumn{2}{c}{\textbf{14B}} & \multicolumn{2}{c}{\textbf{4B}} & \multicolumn{2}{c}{\textbf{12B}} \\
\cmidrule(lr){4-5}\cmidrule(lr){6-7}\cmidrule(lr){8-9}\cmidrule(lr){10-11}
\textbf{Level} & \textbf{Metric} & \textbf{Scoring} &
\textbf{ZS} & \textbf{SFT} &
\textbf{ZS} & \textbf{SFT} &
\textbf{ZS} & \textbf{SFT} &
\textbf{ZS} & \textbf{SFT} \\
\midrule
Text similarity
& ROUGE-1 
& Deterministic
& 30.6 & 49.8 & 12.9 & 50.5 & 25.6 & 46.9 & 30.0 & 49.4 \\

Gold-history turn
& All-turn success 
& LLM judge
& 29.2 & 50.7 & 37.1 & 54.6 & 20.1 & 41.1 & 27.5 & 44.5 \\

Gold-history turn
& Text-turn success 
& LLM judge
& 32.4 & 61.4 & 41.8 & 64.4 & 24.7 & 50.8 & 33.8 & 56.9 \\

Gold-history turn
& Tool-turn success 
& LLM judge
& 21.7 & 26.5 & 26.5 & 32.5 & 9.6 & 19.3 & 13.3 & 16.3 \\

Strict tool check
& Exact call accuracy 
& Deterministic
& 3.6 & 13.9 & 5.4 & 18.1 & 3.0 & 0.0 & 1.8 & 1.8 \\

Strict tool check
& Argument F1 
& Deterministic
& 6.3 & 24.8 & 10.7 & 29.0 & 5.1 & 0.5 & 4.5 & 2.9 \\

Closed-loop replay
& Tool-workflow completion 
& Deterministic
& 0/77 & 3/77 & 0/77 & 8/77 & 0/77 & 0/77 & 0/77 & 0/77 \\

Holistic judge
& Tool-workflow success 
& LLM judge
& 0/77 & 0/77 & 0/77 & 0/77 & 0/77 & 0/77 & 0/77 & 0/77 \\
\bottomrule
\end{tabular}
\end{adjustbox}
\caption{Results across the evaluation protocols for Qwen3 (4B/14B) and Gemma 3 (4B/12B) under Zero-Shot (ZS) and Supervised Fine-Tuning (SFT) settings.}
\label{tab:ladder}
\end{table*}

\section{Results}
\label{sec:results}

Table~\ref{tab:ladder} reports performance across the evaluation protocol, from gold-history turn-level evaluation to end-to-end workflow execution. Overall, SFT substantially improves text-based responses, but these gains transfer weakly to tool execution and rarely yield completed workflows.

\subsection{SFT improves text-based turn-level performance}
\label{sec:gold-history-results}

Under gold-history evaluation, SFT improves performance across all model families and scales. Averaged across the four models, ROUGE-1 increases by 24.4 points, from 24.8 to 49.2, while LLM-judged text-turn success rises by 25.2 points, from 33.2\% to 58.4\%. Consequently, all-turn success improves by 19.3 points, from 28.5\% to 47.7\%. These consistent gains show that SFT helps models better predict the expected next response when given the correct interaction history.

\subsection{Turn-level gains fail to extend to tool use}
\label{sec:tool-results}

Tool-specific improvements are substantially smaller. LLM-judged tool-turn success increases by only 5.9 points on average, from 17.8\% to 23.7\%, compared with 25.2 points for text turns. Exact call accuracy increases from 3.5\% to 8.5\%, and argument F1 from 6.7\% to 14.3\%. These gains are concentrated in Qwen3-4B and Qwen3-14B. Both Gemma 3 models show little or negative improvement; for example, Gemma 3-4B declines from 3.0\% to 0.0\% in exact call accuracy and from 5.1\% to 0.5\% in argument F1.
Thus, turn-level scores based on gold history can obscure poor tool use: frequent text turns improve strongly after SFT and raise the overall score even when exact tool execution remains poor or deteriorates.

\subsection{Gold-history overestimates workflow success}
\label{sec:workflow-results}

The gap widens when models execute workflows using their own prior outputs. Before SFT, no model completes a tool-requiring workflow. After SFT, Qwen3-4B and Qwen3-14B complete only 3 and 8 workflows, respectively, while neither Gemma 3 model completes any. The best result, from Qwen3-14B, is only 10.4\% workflow completion.
Moreover, deterministic completion does not imply overall interaction success. The holistic judge identifies no successful tool-requiring workflow for any model, before or after SFT. A workflow may complete its tool calls while still failing due to missing information, incorrect responses, policy violations, or improper use of tool results.

\subsection{Evaluation protocol changes conclusion}
\label{sec:protocol-gap}

The two protocols yield different conclusions about the same SFT interventions. Gold-history evaluation indicates broad improvement: every model achieves higher ROUGE-1, text-turn success, and all-turn success after SFT. End-to-end execution instead shows that nearly all tool-requiring workflows still fail, with none succeeding under holistic evaluation.
This discrepancy arises because gold-history evaluation restores the correct state before each decision. It tests whether a model can predict the next action from a correct history, but not whether it can construct and maintain that history through its own decisions. End-to-end execution exposes this limitation by allowing early errors to propagate.
Turn-level evaluation is therefore informative but measures a narrower capability than workflow execution. Reliable agent evaluation should report text quality, tool correctness, workflow completion, and holistic trajectory success separately, rather than treating aggregate next-turn performance as evidence of end-to-end capability.

\section{Conclusion}

Across four model pairs from two families, SFT clearly improves reference-matched text and next-action prediction under gold-history evaluation. However, tool-call correctness varies across metrics, model sizes, and families, while end-to-end completion remains low as holistic evaluation finds no successful tool-requiring workflow at any scale. These results show that strong next-turn performance does not necessarily indicate that an agent can complete a workflow using its own interaction history. Future work should validate this finding across additional domains, model families, and interactive environments, while developing training methods that explicitly target error recovery, state maintenance, and end-to-end task completion. More broadly, agent evaluations should combine turn-level metrics with tool correctness and workflow-level success to provide a more complete account of agent capability.


\section*{Limitations}

Our evaluation covers two model families and one proprietary customer-support domain, limiting generalizability. Closed-loop replay requires exact reference tool calls and uses deterministic gold user turns, potentially rejecting valid alternatives and underrepresenting real interactions. 

\section*{Ethics Statement}

All conversations are synthetic and contain no customer data or personally identifiable information. To help facilitate future work, sanitized prompt templates are provided in the Appendix (see Section \ref{app:data-prompts} and Section \ref{app:judge-prompts}).  

\bibliography{custom}

\appendix
\label{appendix}

\section{Sanitized Data-Generation Prompts}
\label{app:data-prompts}

The following prompts preserve the roles, inputs, and decisions used in the
pipeline while replacing proprietary policies, tools, and company information
with placeholders.

\subsection{Scenario Generation}

\begin{Verbatim}[fontsize=\scriptsize,frame=single]
Given the domain policy and available tools, create a
realistic customer-support scenario for this workflow.

Specify:
- the user's objective and relevant initial state;
- the desired final state;
- information the user initially knows;
- the expected tool operations; and
- observable criteria for successful completion.

The scenario must be solvable using only the policy and
tools supplied below. Use synthetic identities and data.

DOMAIN POLICY:
<SANITIZED_DOMAIN_POLICY>

AVAILABLE TOOLS:
<SANITIZED_TOOL_DEFINITIONS>
\end{Verbatim}

\subsection{GPT-5 User Simulator}

\begin{Verbatim}[fontsize=\scriptsize,frame=single]
Act as the customer described below. Remain in character
and pursue the assigned objective naturally.

- Reveal information gradually and only when appropriate.
- Answer the agent's questions consistently with the
  scenario state and persona.
- Do not mention tools, policies, prompts, or simulation.
- End when the goal is met or cannot be completed.

SCENARIO AND OBJECTIVE:
<SANITIZED_SCENARIO>

SYNTHETIC IDENTITY AND PERSONA:
<SYNTHETIC_USER_PROFILE>

CONVERSATION SO FAR:
<DIALOGUE_HISTORY>

Return only the next user message.
\end{Verbatim}

\subsection{GPT-5 Workflow Agent}

\begin{Verbatim}[fontsize=\scriptsize,frame=single]
Act as a customer-support agent operating under the
domain policy below.

- Follow the workflow and all policy constraints.
- Maintain the conversation state and request missing
  information when needed.
- Call an available tool when the workflow requires it.
- Use tool results before making state-dependent claims.
- Handle invalid requests and tool errors safely.

Return either the next natural-language response or a
structured tool call.

DOMAIN POLICY:
<SANITIZED_DOMAIN_POLICY>

AVAILABLE TOOLS:
<SANITIZED_TOOL_DEFINITIONS>

CONVERSATION SO FAR:
<DIALOGUE_HISTORY_WITH_TOOL_RESULTS>
\end{Verbatim}

\subsection{Conversation Validation}

This prompt is run independently with each validation judge. A conversation is
retained only when both judges approve it.

\begin{Verbatim}[fontsize=\scriptsize,frame=single]
Review the complete synthetic conversation for use as a
workflow-training example.

Approve it only if the agent follows the supplied policy,
uses tools appropriately, remains consistent with tool
results, and reaches a sensible outcome for the scenario.
Reject it for a missing required step, contradiction,
unsupported claim, invalid tool use, policy violation, or
inconsistent simulated-user behavior.

DOMAIN POLICY AND TOOLS:
<SANITIZED_POLICY_AND_TOOLS>

SCENARIO:
<SANITIZED_SCENARIO>

GENERATED CONVERSATION:
<SYNTHETIC_CONVERSATION>

Return JSON only:
{
  "decision": "APPROVE or REJECT",
  "reasons": ["brief reason"]
}
\end{Verbatim}

\section{Sanitized Evaluation-Judge Prompts}
\label{app:judge-prompts}

These prompts show the information supplied to the judges and the binary
decisions reported in the paper. Auxiliary diagnostic fields used during
analysis are omitted.

\subsection{Gold-History Turn Judge}

\begin{Verbatim}[fontsize=\scriptsize,frame=single]
Evaluate one assistant decision in a multi-turn
customer-support workflow. Judge only what should happen
at this turn; do not require final workflow completion at
an intermediate turn.

Use the policy, tools, correct history, and reference
action to determine whether the prediction is an
acceptable next action. Alternative wording and safe,
semantically equivalent actions are allowed. A tool call
must select an appropriate function and use arguments
grounded in the supplied history.

DOMAIN POLICY:
<SANITIZED_DOMAIN_POLICY>

AVAILABLE TOOLS:
<SANITIZED_TOOL_DEFINITIONS>

CORRECT HISTORY BEFORE THIS TURN:
<GOLD_HISTORY>

REFERENCE ACTION:
<REFERENCE_ACTION>

MODEL PREDICTION:
<MODEL_OUTPUT>

Return JSON only:
{
  "turn_success": true or false,
  "failure_category": "none or primary failure",
  "rationale": "brief evidence-based explanation"
}
\end{Verbatim}

\subsection{Holistic Workflow Judge}

\begin{Verbatim}[fontsize=\scriptsize,frame=single]
Evaluate whether the candidate replay completes the
user's workflow under the supplied domain policy.

Use the reference trajectory to understand the scenario,
required state changes, and acceptable final outcome. Do
not reward wording similarity. Safe alternative wording
or an equivalent valid path is allowed, but required
verification, policy constraints, and consequential tool
actions must be satisfied.

Accepted tool attempts received the recorded tool result.
Rejected attempts were invalid at that workflow state. A
replay that ends before the user goal is resolved is not
successful.

DOMAIN POLICY AND TOOLS:
<SANITIZED_POLICY_AND_TOOLS>

REFERENCE TRAJECTORY:
<SANITIZED_REFERENCE_TRAJECTORY>

CANDIDATE CLOSED-LOOP REPLAY:
<SANITIZED_CANDIDATE_REPLAY>

Return JSON only:
{
  "success": true or false,
  "failure_category": "none or primary failure",
  "rationale": "brief evidence-based explanation"
}
\end{Verbatim}

\end{document}